\documentclass[]{elsarticle}

\usepackage{enumitem}
\usepackage[]{natbib}
\usepackage{graphicx} 
\usepackage{subfiles}
\usepackage{multirow}
\usepackage{geometry}
\usepackage{url}
\usepackage{hyperref}
\usepackage{booktabs}
\usepackage{verbatim}
\usepackage{amsmath}
\usepackage{xcolor}
\usepackage{subfig}
\usepackage{listings}
\usepackage{tcolorbox}
\usepackage{tabularx}

\newif\ifsubfile
\subfiletrue

\title{Automated Identification of Incidentalomas Requiring Follow-Up: A Multi-Anatomy Evaluation of LLM-Based and Supervised Approaches}

\author{Namu Park\textsuperscript{1},
Farzad Ahmed\textsuperscript{2},
Zhaoyi Sun\footnotemark[1],
Kevin Lybarger\footnotemark[2],
Ethan Breinhorst\footnotemark[3],
Julie Hu\footnotemark[3],
Özlem Uzuner\footnotemark[2],
Martin Gunn\footnotemark[4],
Meliha Yetisgen\footnotemark[1]}

\address{
$^{1}$Department of Biomedical Informatics and Medical Education, University of Washington, Seattle, WA, USA\\
$^{2}$Department of Information Sciences and Technology, George Mason University, Fairfax, VA, USA\\
$^{3}$Department of Radiology, Te Whatu Ora Health New Zealand, Te Toka Tumai Auckland, Auckland, New Zealand\\
$^{4}$Department of Radiology, School of Medicine, University of Washington, Seattle, WA, USA\\[4pt]
}

\begin{document}

\begin{abstract}

\noindent \textbf{Objective:} To evaluate large language models (LLMs) against supervised baselines for fine-grained, lesion-level detection of incidentalomas requiring follow-up, addressing the limitations of current document-level classification systems.

\vspace{0.5em}
\noindent \textbf{Methods:} We utilized a dataset of 400 annotated radiology reports containing 1,623 verified lesion findings. We compared three supervised transformer-based encoders (BioClinicalModernBERT, ModernBERT, Clinical Longformer) against four generative LLM configurations (Llama 3.1-8B, GPT-4o, GPT-OSS-20b). We introduced a novel inference strategy using lesion-tagged inputs and anatomy-aware prompting to ground model reasoning. Performance was evaluated using class-specific F1-scores.

\vspace{0.5em}
\noindent \textbf{Results:} The anatomy-informed GPT-OSS-20b model achieved the highest performance, yielding an incidentaloma-positive macro-F1 of 0.79. This surpassed all supervised baselines (maximum macro-F1: 0.70) and closely matched the inter-annotator agreement of 0.76. Explicit anatomical grounding yielded statistically significant performance gains across GPT-based models ($p < 0.05$), while a majority-vote ensemble of the top systems further improved the macro-F1 to 0.90. Error analysis revealed that anatomy-aware LLMs demonstrated superior contextual reasoning in distinguishing actionable findings from benign lesions.

\vspace{0.5em}
\noindent \textbf{Conclusion:} Generative LLMs, when enhanced with structured lesion tagging and anatomical context, significantly outperform traditional supervised encoders and achieve performance comparable to human experts. This approach offers a reliable, interpretable pathway for automated incidental finding surveillance in radiology workflows.

\end{abstract}


\maketitle

\subfilefalse

\section{Introduction}

Incidental findings, or \textit{incidentalomas}, refer to unexpected abnormalities discovered during imaging studies performed for unrelated reasons \cite{berland2010managing}. Their detection has increased as imaging utilization has grown across healthcare. A systematic review estimated that incidental findings appear in up to one-third of computed tomography (CT) and positron emission tomography CT (PET-CT) examinations \cite{lumbreras2010incidental}. These findings create a clinical dilemma, since most are benign while some represent early-stage disease that requires intervention. Balancing overdiagnosis and missed pathology has become an important concern in medical practice.

To support clinicians, the American College of Radiology (ACR) and other professional groups provide standardized decision rules for managing incidental findings \cite{berland2010managing}. These guidelines outline follow-up pathways based on organ type, lesion size, and imaging features. Ensuring adherence remains difficult in real-world settings. Radiology reports are unstructured narratives that often include multiple findings and subtle recommendations, so significant incidental findings may be overlooked or receive no follow-up, which contributes to patient risk and healthcare inefficiency. Decision rules also contain little clinical context about comorbidities, although this information is relevant for determining the appropriate timing and type of follow-up.

In large hospital systems, millions of radiology reports are generated each year, so manual review is infeasible. Automated systems that identify incidental findings and follow-up recommendations could improve care coordination. Early natural language processing (NLP) work demonstrated that extracting such information is possible \cite{cai2016nlp, mabotuwana2018improving, dalal_determining_2020}, but earlier systems were limited by linguistic variability, contextual ambiguity, and the lack of lesion-level supervision. Distinguishing an incidental lesion from the primary diagnostic concern often requires nuanced understanding that rule-based or shallow learning approaches cannot provide.

Recent advances in clinical NLP offer new opportunities. Domain-specific encoders such as BioClinicalBERT \cite{alsentzer2019publicly} and ClinicalBERT \cite{huang2019clinicalbert} introduced contextual representations tailored to clinical text. Generative models such as GPT-4 \cite{nori2023capabilities} and Med-PaLM \cite{singhal2023nature} show strong reasoning abilities and can interpret subtle contextual cues, recognize implicit recommendations, and integrate cross-sentence information that is essential for identifying incidentalomas accurately.

This study uses improved reasoning and text-understanding capabilities of modern NLP systems to develop LLM-based methods for anatomy-aware identification of incidentalomas. By integrating lesion-tagged inputs and prompting strategies grounded in anatomical structure, our approach aligns computational predictions with how radiologists interpret imaging findings. We compare LLM-based systems with strong supervised baselines to evaluate their accuracy and clinical consistency. We find that anatomical grounding improves performance and interpretability. The best LLM configuration achieves an incidentaloma-positive macro-F1 of 0.79, which closely matches the inter-annotator agreement macro-F1 of 0.76.

\begin{table}[ht!] \centering \renewcommand{\arraystretch}{1.4} \caption*{\textbf{Statement of significance}} \begin{tabular}{|p{0.23\linewidth}|p{0.70\linewidth}|} \hline \textbf{Problem} & Radiology reports frequently contain incidental findings, yet many are under-recognized or lack appropriate follow-up, leading to missed or delayed diagnoses. Manual review is infeasible at scale, and existing automated systems often fail to distinguish incidentalomas from clinically indicated findings. \\ \hline \textbf{What is Already Known} & Previous NLP and machine learning methods have demonstrated the feasibility of extracting incidental findings from radiology text but mostly rely on document-level classification. These systems struggle with contextual ambiguity and lack lesion-level interpretability. \\ \hline \textbf{What this Paper Adds} & This paper presents a lesion-tagged, anatomy-aware framework for automated identification of incidentalomas requiring follow-up across six anatomies. By integrating structured lesion information with LLMs and supervised encoders, the study shows that anatomy-informed prompting substantially improves detection accuracy, interpretability, and clinical reliability. It also introduces the first lesion-level benchmark and error analysis for incidentaloma classification. \\ \hline \textbf{Who Would Benefit from the New Knowledge in this Paper} & Radiologists, clinical informaticians, and health systems developing scalable AI tools for incidental finding surveillance and data-driven quality improvement in imaging workflows. \\ \hline \end{tabular} \end{table}

\ifsubfile
\bibliography{mybib}
\fi

\section{Related Work}

\subsection{Clinical NLP using Supervised Models}

Transformer-based language models \cite{vaswani2017attention} have transformed clinical NLP by leveraging contextual embeddings from large-scale biomedical and clinical corpora. BioClinicalBERT \cite{alsentzer2019publicly}, trained on MIMIC-III \cite{johnson2016mimic3} and PubMed texts, has demonstrated strong performance on concept extraction and relation detection tasks. Variants such as ClinicalBERT \cite{huang2019clinicalbert} and other transfer learning approaches have further improved robustness across multiple clinical domains \cite{peng2019transfer}. To handle longer clinical documents, architectures such as Clinical Longformer and Clinical BigBird extend transformers with sparse attention, enabling efficient modeling of radiology reports and other long sequences \cite{li2022clinical}. More recently, domain-extended models like ModernBERT \cite{warner2024smarter}, BioClinicalModernBERT \cite{sounack2025bioclinical} incorporate long-context adaptations specifically designed for biomedical applications.

In radiology, these models have been applied to tasks such as report classification, lesion entity extraction, and malignancy detection \cite{irvin2019chexpert,gerevini2018automatic}. Comparative evaluations highlight their strengths for document-level prediction but also reveal challenges in multi-class classification tasks where false negatives are especially costly \cite{welch2010overdiagnosis,macmahon2017guidelines}. Our work contributes to this body of research by systematically benchmarking BioClinicalModernBERT, ModernBERT, and Clinical Longformer for incidentaloma detection, providing evaluation across multiple anatomies.

\subsection{Clinical NLP using Generative Large Language Models}

Generative LLMs have recently emerged as powerful tools for clinical NLP, capable of zero-shot and few-shot generalization across diverse medical tasks. GPT-4 \cite{openai2023gpt4} and GPT-4o \cite{hurst2024gpt} both have shown considerable performance on medical reasoning benchmarks, demonstrating competency in both multiple-choice exams and free-text reasoning tasks \cite{nori2023capabilities, hurst2024gpt, park2025identifying}. Similarly, Med-PaLM \cite{singhal2023nature}, an instruction-tuned variant of PaLM \cite{chowdhery2023palm}, has exhibited strong performance on clinically-oriented question answering and reasoning tasks. Open-source foundation models such as LLaMA \cite{touvron2023llama} have also been widely adopted in clinical NLP research as the backbone for lightweight adaptations. Methods like low-rank adaptation (LoRA) \cite{hu2022lora} have enabled efficient fine-tuning of LLMs for domain-specific tasks without retraining full models.

Within radiology domain, LLMs have been explored for report summarization \cite{liu25radiology}, clinical reasoning \cite{wang2023clinicalgpt}, and structured entity extraction from imaging reports \cite{park2024novel}. Instruction-tuned medical LLMs such as MedAlpaca \cite{han2023medalpaca} have further demonstrated the feasibility of adapting general-purpose models to healthcare through domain-specific instruction datasets. Despite these advances, few studies have explored lesion-level classification tasks that explicitly leverage lesion and anatomy information. Our work addresses this gap by 1) marking candidate lesions with structured tags in the report text and 2) providing LLMs with explicit lesion-anatomy associations during inference. These structured inputs enhance classification and interpretability by ensuring models attend to specific findings within their correct anatomical context.

\subsection{Identification of Incidentalomas in Radiology Reports}

Large-scale analyses have highlighted the frequency of incidental findings in abdominal and pelvic CT exams and their downstream clinical implications \cite{lumbreras2010incidental}. Organ-specific investigations have characterized incidentalomas of the adrenal glands, thyroid, and lungs, providing prevalence estimates and evidence-based follow-up recommendations \cite{maher2018adrenal,song2024incidental,macmahon2017guidelines}. Traditional approaches have relied on structured radiology databases and manual chart reviews, which ensure accuracy but limit scalability across large health systems \cite{reiner2007radiology,welch2010overdiagnosis}.

Earlier NLP systems established the foundation for automated detection of incidental findings in radiology reports. Dutta \textit{et al.} \cite{dutta2013automated} and Trivedi \textit{et al.} \cite{trivedi2019interactive} implemented rule-based and interactive NLP frameworks for identifying incidental observations in narrative imaging text, while Kang \textit{et al.} \cite{kang2019nlp} focused on incidental pulmonary nodules. These early studies demonstrated the feasibility of text mining for large-scale retrospective identification of incidental findings but were limited by hand-engineered features and domain-specific lexicons. Building on these foundations, more recent NLP pipelines have leveraged machine learning and transformer-based architectures to automatically extract incidental findings from unstructured radiology reports \cite{cai2016nlp,park2024novel}. However, most prior NLP efforts have emphasized document-level classification of reports containing incidental findings rather than fine-grained lesion-level annotation. Closely related to incidentaloma identification, recent studies have advanced the extraction of follow-up imaging recommendations from radiology reports using NLP approaches \cite{mabotuwana2018improving, lau2020extraction, dalal_determining_2020}.
  
Subsequent advances in deep learning enabled more robust classification and extraction pipelines. Canton \textit{et al.} \cite{canton2021automatic} proposed a multi-modal system that combined imaging metadata and report text to automatically detect adrenal and thyroid incidentalomas. Schumm \textit{et al.} \cite{schumm2023adrenal} automated adrenal nodule extraction from electronic health records, and Bala \textit{et al.} \cite{bala2020web} developed a web-based application for incidentaloma tracking and management, illustrating the integration of AI systems into clinical follow-up workflows.  

Recent work has highlighted the growing role of LLMs in incidental finding interpretation and communication. Woo \textit{et al.} \cite{woo2025gpt4incidental} evaluated a HIPAA-compliant instance of GPT-4 \cite{openai2023gpt4} for identifying actionable incidental findings and generating patient-facing summaries from radiology reports, achieving high factual alignment with expert reviews. Bhayana \textit{et al.} \cite{bhayana2024use} similarly demonstrated that single-shot prompting with GPT-4 could identify incidental findings across multiple report types with minimal supervision. Aksu \textit{et al.} \cite{aksu2025aiendocrine} compared ChatGPT \cite{openai2023gpt4}, Gemini \cite{team2023gemini}, and Claude \cite{anthropic2024claude} for adrenal incidentaloma management, showing that GPT-based reasoning systems can approximate endocrinologist-level decision-making. These developments highlight the potential of multimodal and text-based LLM systems to enhance incidentaloma triage and follow-up workflows. Our study extends this line of work by introducing lesion-specific annotations across six anatomies and explicitly modeling both incidentaloma presence and follow-up requirements, while exploring how LLM-based prompting can improve lesion-level interpretability and clinical decision support.

\ifsubfile
\bibliography{mybib}
\fi

\section{Methods}

\subsection{Dataset}

\subsubsection{Preprocessing}
\label{sec:preprocessing}

We utilized an existing clinical database comprising 6,668,323 radiology reports from 2007 to 2020, representing the general patient population across four hospitals within the UW Medicine system. All reports were automatically de-identified and processed with \textit{PL-Marker++}, a radiology-specific BERT-based model for lesion-level entity and relation extraction \cite{leeDobbins2021, park2024novel}. PL-Marker++ was trained on a manually annotated corpus of 609 radiology reports labeled by medical residents and board-certified radiologists, achieving inter-annotator agreement F1 = 0.762 and macro-F1 = 0.88 for lesion finding extraction.

PL-Marker++ extracts three main categories of information: clinical indications, lesion findings, and medical problems, each accompanied by finding-specific arguments such as anatomy and size trend attributes. Clinical indications describe the reason for the imaging study (e.g., \textit{``evaluation of lung nodule'', ``follow-up for prior mass'',} or \textit{``abdominal pain''}) and are automatically categorized into four subtypes: \textit{neoplastic diagnosis}, \textit{non-neoplastic diagnosis}, \textit{symptom}, and \textit{trauma}. Lesion findings represent mass-forming pathologic structures (e.g., \textit{``hepatic lesion'', ``adrenal nodule''}), whereas medical problems represent non mass-forming pathologic processes. Anatomical and descriptive attributes capture details such as size, count, and temporal size trend with five possible values: \textit{increasing}, \textit{decreasing}, \textit{no change}, \textit{disappeared}, and \textit{new}. Anatomy information associated with each lesion, medical problem, and clinical indication is extracted using entity–relation extraction and mapped to predefined anatomy categories such as \textit{lung}, \textit{liver}, and \textit{kidney}. 

Table~\ref{tab:lesion_identification} presents token-level lesion identification performance on the annotated dataset described by Park et al.\ \cite{park2024novel}. On the test set, PL-Marker++, a BERT-based model trained on annotated radiology reports, outperformed $n$-shot approaches using Llama 3.1-8B and GPT-4o. This is consistent with recent work \cite{hu2024improving, lu2025large} showing that LLMs often struggle with token-level clinical named entity recognition tasks, likely due to limitations of decoder-based architectures. 

These structured outputs, particularly the integration of categorized clinical indications and lesion-level findings with anatomical and size trend attributes, formed the basis for this study. They enabled the identification of reports describing new or potentially actionable lesions and provided clinical context to improve LLM inference performance described in the following sections. The study protocol was reviewed and approved by the University of Washington Institutional Review Board (IRB).

\begin{table}[]
\centering
\caption{Lesion identification performance using multiple approaches on annotated radiology reports (Park et al., 2024).}
\label{tab:lesion_identification}
\begin{tabular}{l|cccccc}
\hline
\multicolumn{1}{c|}{\multirow{2}{*}{Method}} & \multicolumn{3}{c|}{Overlap Match}                 & \multicolumn{3}{c}{Exact Match} \\ \cline{2-7} 
\multicolumn{1}{c|}{}                        & Precision & Recall & \multicolumn{1}{c|}{F1-score} & Precision  & Recall  & F1-score \\ \hline
PL-Marker ++           & 0.880 & 0.888 & \textbf{0.884} & 0.740 & 0.712 & \textbf{0.726} \\
Llama 3.1-8B (0-shot) & 0.377 & 0.322 & 0.348 & 0.014 & 0.012 & 0.013 \\
Llama 3.1-8B (1-shot) & 0.449 & 0.387 & 0.415 & 0.064 & 0.055 & 0.059 \\
GPT-4o (0-shot)          & 0.599 & 0.374 & 0.460 & 0.170 & 0.106 & 0.131 \\
GPT-4o (1-shot)          & 0.576 & 0.489 & 0.529 & 0.192 & 0.163 & 0.177 \\ \hline
\end{tabular}
\end{table}

\subsubsection{Sampling Process}\label{sec:sampling}

The low prevalence of incidentaloma-containing reports (e.g. 9.9\% for adrenal incidentalomas \cite{bala2020web} and less than 5\% for chest computed tomography for incidental pulmonary embolism \cite{o2018prevalence}) and the substantial linguistic variability with which radiologists describe these findings make identifying such cases challenging. Directly searching for terms such as “incidental” or “incidentaloma” would introduce bias and miss many true incidentalomas, since radiologists often describe them without using these terms. To avoid this limitation, we used a broader sampling strategy that leveraged PL-Marker++ outputs and SVM-based recommendation sentence identification, aiming for more comprehensive coverage of incidentaloma cases. After discussions with a board-certified radiologist, we developed a sampling strategy designed to achieve a high positivity rate for reports likely to contain incidentalomas. Although this approach may introduce some sampling bias, it increased the proportion of incidentaloma-positive reports, which was important for constructing a robust gold standard. Our sampling strategy applied four sequential filters:

\begin{enumerate}[label=(\arabic*)]
    \item \textbf{Target relevant anatomies:} Using structured clinical findings extracted with PL-Marker++, we analyzed all radiology reports and identified findings mapped to six anatomical regions: kidney, liver, lung, pancreas, adrenal gland, and thyroid. These regions were selected because they are more likely to contain incidentalomas. Among all reports, 24.7 \% (n = 1,767,623) contained at least one finding (clinical indication, medical problem, or lesion) from these anatomies, yielding 7,519,138 findings. We then selected reports across all imaging modalities that included lesion findings in these regions with assertion values labeled as \textit{``present''} or \textit{``possible''}. In total, 6.3 \% of all reports (n = 112,100) satisfied these criteria and were retained.
    \item \textbf{Exclude previously identified lesions:} Using size trend data, we excluded reports with size trend values of ``\textit{increasing}'', ``\textit{decreasing}'', ``\textit{disappeared}'', or ``\textit{no change}'', which indicate comparison with prior imaging and therefore previously identified lesions. Accordingly, 5.7\% of reports identified in step (1) were excluded, leaving 105,729 reports.
    \item \textbf{Filter surveillance cases:} PL-Marker++ was used to extract clinical indication types (\textit{trauma}, \textit{symptom}, \textit{neoplastic diagnosis}, and \textit{non-neoplastic diagnosis}). Reports with a neoplastic diagnosis in the clinical indication were excluded to minimize inclusion of expected findings rather than incidental ones. After this step, 39.6\% of reports (n = 41,833) were retained.
    \item \textbf{Select reports with follow-up recommendations:} Among the 41,833 reports retained, we further selected those containing follow-up recommendation sentences, as incidentalomas are frequently accompanied by such recommendations. Recommendation sentences were identified using a support vector machine (SVM)–based classifier \cite{lau2020extraction}. This step yielded 19,690 reports with the highest probability of containing a clinically important incidentaloma.
\end{enumerate}

To ensure accurate anatomical assignment for each lesion and facilitate lesion-level and document-level analyses, we implemented a double-verification process for anatomy extraction. PL-Marker++ initially provided sentence-level lesion–anatomy mappings, but its inference was sometimes limited when relevant anatomy mentions appeared outside the immediate sentence. To address this, we used an additional LLM (Llama 3.1-8B-Instruct) \cite{grattafiori2024llama} trained to infer anatomical associations from full report context, which achieved a micro-F1 score of 0.895 on the annotated dataset from the PL-Marker++ study. Discrepancies between systems were manually reviewed, and a final verified anatomy label was assigned to each annotated lesion. This dual-level verification ensured robust and consistent anatomical categorization across six target anatomies: kidney, liver, lung, pancreas, adrenal gland, and thyroid.

\subsubsection{Data Annotation} \label{sec:data_annotation}
Of the 19,690 radiology reports that met the selection criteria, 400 reports were randomly selected and annotated. Annotation guidelines were refined through multiple iterations by two board-certified radiologists to ensure both precision and clinical relevance. Rather than relying solely on document-level labels, the guidelines were designed to leverage lesion-level findings. Each lesion was annotated as \textit{No Incidentaloma}, \textit{Incidentaloma--No Risk}, or \textit{Incidentaloma--Follow-up Required}, with the distinction based on clinical severity (e.g. \textit{``Benign simple cysts''} were considered as not requiring follow-up). To expedite annotation and minimize missed lesion mentions, lesion findings extracted using PL-Marker++ \cite{park2024novel} were pre-annotated in the BRAT annotation tool \cite{stenetorp2012brat}. Differential diagnoses were excluded.

Two medical residents participated in the annotation process. Initially, we conducted double-annotation training rounds in which both annotators independently labeled the same reports, followed by weekly feedback sessions supervised by a board-certified radiologist. Discrepancies were reviewed and resolved through consensus, and remaining disagreements were adjudicated by the radiologist who developed the guidelines. Once a document-level inter-annotator agreement (IAA) of 0.896 F1 for incidentaloma status was achieved, we transitioned to single-annotation rounds, where each annotator was assigned distinct reports.

As a result, 160 double-annotated reports contained 1,623 pre-identified lesion findings, averaging 10.15 lesions per report. Of these reports, 117 (73.1\%) contained at least one incidentaloma, demonstrating an enriched sample. Annotators reached agreement on 1,498 incidentaloma labels, resulting in an agreement rate of 92.3\% (0.838 F1). Common incidentaloma terms included ``lesion'', ``nodule'', ``cyst'', and ``mass''. Annotator 1 identified an average of 2.538 incidentalomas per report, slightly higher than Annotator 2’s average of 2.294. We then moved to single annotation for the remaining reports. The overall dataset distribution at the document level is summarized in Table \ref{tab:tab1}, while Table \ref{tab:tab2} presents the anatomy-specific distribution of all annotated lesion findings.

\begin{table}[ht]
\centering
\caption{Distribution of annotated reports with and without incidentalomas.}
\label{tab:tab1}
\begin{tabular}{lccc}
\hline
 & Total (n) & W/ Incidentaloma & W/O Incidentaloma \\
\midrule
Double-annotated & 160 & 117 (73.1\%) & 43 (26.9\%) \\
Single-annotated & 240 & 158 (65.8\%) & 82 (34.2\%) \\
\midrule
Total & 400 & 275 (69.25\%) & 125 (30.75\%) \\
\hline
\end{tabular}
\end{table}

\subsection{Incidentaloma Classification}\label{sec:classification}

Using the annotated reports, we defined the incidentaloma classification task to support both lesion-level and document-level analyses. Leveraging anatomy information from Section \ref{sec:sampling} together with the annotated lesion findings, we framed incidentaloma identification as seven anatomy-specific, three-way classifications per report. For each anatomy (\textit{Lung, Liver, Kidney, Adrenal, Pancreas, Thyroid, Other}), the model generated one label from \{0 = \textit{No Incidentaloma}, 1 = \textit{Incidentaloma--No Risk}, 2 = \textit{Incidentaloma--Follow-up Required}\}.

During inference, all models first generated lesion-level predictions for each anatomy and then aggregated these predictions into a single anatomy-level label using a severity precedence rule. Formally, for a given anatomy $a$ with lesion labels $\{l_1, l_2, \dots, l_n\}$, the aggregated label $L_a$ was defined as:

\[
L_a = \max \{ l_1, l_2, \dots, l_n \}, \quad l_i \in \{0,1,2\},
\]

where $0 =$ No Incidentaloma, $1 =$ Incidentaloma--No Risk, and $2 =$ Incidentaloma--Follow-up Required. For example, if three lung lesions were labeled $\{0, 0, 2\}$, the aggregated anatomy label for lung would be $L_{\text{lung}} = 2$. These class indices ($0$, $1$, and $2$) are used throughout the paper to refer to their corresponding labels.

Each radiology report was represented as a structured vector of seven anatomy-specific incidentaloma labels:

\[
R = (L_{\text{lung}}, L_{\text{liver}}, L_{\text{kidney}}, 
     L_{\text{adrenal}}, L_{\text{pancreas}}, 
     L_{\text{thyroid}}, L_{\text{other}}), \quad L_a \in \{0,1,2\}.
\]

Inter-annotator agreement (IAA) for the document-level annotations, calculated without considering anatomy-specific labels, was 0.93 F1 for \textit{No Incidentaloma}, 0.81 F1 for \textit{Incidentaloma--No Risk}, and 0.70 F1 for \textit{Incidentaloma--Follow-up Required}, resulting in a macro-average F1 of 0.81. This schema provided consistent, anatomy-specific labeling and served as the foundation for subsequent model training and evaluation. Summary statistics are presented in Table~\ref{tab:tab2}.

\begin{table}[]
\centering
\caption{Number of incidentalomas across six target anatomies in our dataset of 400 radiology reports. A single report may include multiple incidentalomas in different anatomical sites. Percentages in parentheses indicate the proportion of low-risk incidentalomas and those requiring follow-up within each anatomy.}
\label{tab:tab2}
\begin{tabular}{lccc}
\hline
\multirow{2}{*}{} & \multicolumn{1}{l}{\multirow{2}{*}{No Incidentaloma}} & \multicolumn{2}{c}{Incidentaloma}                                              \\ \cline{3-4} 
                  & \multicolumn{1}{l}{}                                  & \multicolumn{1}{c}{No Risk} & \multicolumn{1}{c}{Follow-up Required} \\ \hline
Lung     & 324 & 18 (4.5\%)  & 58 (14.5\%)\\
Liver    & 302 & 78 (19.5\%) & 20 (5.0\%) \\
Kidney   & 238 & 128 (32.0\%) & 34 (8.5\%) \\
Adrenal  & 379 & 6 (1.5\%)  & 15 (3.8\%) \\
Pancreas & 384 & 5 (1.2\%)   & 11 (2.8\%) \\
Thyroid  & 378 & 14 (3.5\%)  & 8 (2.0\%)  \\
Others   & 341 & 25 (6.3\%)  & 34 (8.5\%) \\ \hline
Total & 2346 & 274 & 180 \\ \hline
\end{tabular}
\end{table}

\subsubsection{Supervised Learning Approach}

We fine-tuned three transformer-based encoders on the annotated dataset: BioClinicalModernBERT, ModernBERT, and Clinical Longformer. BioClinicalModernBERT \cite{sounack2025bioclinical} is a recent BERT-based model for bioclinical applications pre-trained on large-scale biomedical and clinical corpora, optimized for domain-specific terminology and extended context modeling. ModernBERT \cite{warner2024smarter} combines general-domain and clinical text during pre-training, providing a balanced representation that enables comparison with a less domain-specialized encoder. Clinical Longformer \cite{li2022clinical} uses a sparse attention mechanism to efficiently process lengthy radiology reports that often exceed the input capacity of standard BERT-like models. All models were trained according to the classification task described in Section~\ref{sec:classification}, to extract lesion-level incidentaloma labels.

Model hyperparameters were tuned on the validation set. We explored learning rates in $\{1\times10^{-5}, 2\times10^{-5}, 3\times10^{-5}, 5\times10^{-5}\}$, batch sizes of $\{8, 16, 32\}$, and up to 15 training epochs, with weight decay $\{0.01, 0.05\}$ and dropout $\{0.1, 0.2\}$. The final configuration, selected according to the highest validation macro-F1 for incidentaloma classes, used a batch size of 16, learning rate $2\times10^{-5}$, linear warmup over 10\% of total steps, weight decay 0.01, and dropout 0.1. The maximum sequence length was 512 tokens for BioClinicalModernBERT and ModernBERT, and 2{,}048 tokens for Clinical Longformer. Models were trained for up to 10 epochs with early stopping (patience = 3), using the AdamW optimizer \cite{loshchilov2017decoupled}, gradient clipping \cite{zhang2019gradient} (maximum norm = 1.0), and mixed-precision (FP16) training on NVIDIA A100 GPUs.

The input features comprised a structured textual representation derived from the preprocessed radiology reports rather than the full report narrative. For each report, we generated seven anatomy-specific input strings, one for each anatomy category. Each input took the form \texttt{Anatomy: <organ> | Lesion: <lesion description> | Context: <neighboring text>}, allowing the model to reason separately about potential incidental findings in each anatomical region. Anatomical information was obtained from the verified anatomy labels generated during preprocessing (Section~\ref{sec:preprocessing}) and inserted directly into the input as plain text. Lesion descriptions were obtained from the lesion spans identified during the same stage, and a maximum of 100 characters of surrounding report text were included to capture relevant local context. This window size was selected through validation-set analysis to balance lesion specificity with contextual information.

Each sequence was tokenized and padded to the maximum input length supported by the encoder model, using the model-specific tokenizer. This yielded a unified text representation capturing three components, the anatomy of interest, the lesion description, and the immediate local context, while ensuring consistent formatting across models with different sequence capacities. During inference, the supervised models first generated lesion-level predictions for each anatomy and then aggregated these into final document-level labels, as described in Section~\ref{sec:classification}.


Since the gold standard labels were substantially imbalanced (Table \ref{tab:tab2}), with relatively few instances labeled \textit{Incidentaloma–Follow-up Required}, we implemented cost-sensitive learning strategies to enhance performance on underrepresented yet clinically important classes. Three approaches were evaluated. First, class-weighted cross-entropy was applied, assigning weights inversely proportional to class frequencies. Second, focal loss was employed with $\gamma=2$ and class-specific weighting $\alpha=w_c$, emphasizing hard-to-classify and minority examples. Finally, an expected-cost objective was tested using a $3\times3$ asymmetric cost matrix $C$ designed to penalize clinically consequential misclassifications:
\[
\mathcal{L}_{\text{EC}} = \frac{1}{N}\sum_{i=1}^{N}\sum_{k=0}^{2} C_{y_i,k}\, p_\theta(k \mid x_i),
\]
where $p_\theta$ denotes the softmax output probabilities. At inference time, we also used cost-aware decision making by selecting the label that minimized expected misclassification cost under the predicted distribution.

\subsubsection{LLM-based Approach}

Similar to supervised learning methods, the LLM-based approach first extracts lesion-level incidentaloma labels and then aggregates these predictions into document-level labels, as described in Section~\ref{sec:classification}. For LLMs, prior to inference, candidate lesion entity spans in the original report were enclosed with XML-format tags using lesion information extracted by PL-Marker++, while the surrounding text remained intact. Numbered lesion tags support detailed error analysis and help reduce noise. Without tags, it can be difficult to identify the lesion of interest, particularly when terms like ``nodule'' appear in multiple contexts within the same report. By highlighting lesions directly, tags disambiguate such cases, expedite manual review by clinicians, and minimize the influence of irrelevant or distracting text, which improves interpretability and clinical reliability.

Our ablation study using the Llama 3.1-8B Instruct model showed that the inclusion of lesion tags substantially improved performance compared with untagged inputs. In the 0-shot setting, inputs with lesion tags achieved higher macro F1-score than all 0-shot, 1-shot and 5-shot settings without tags, indicating that structural cues can be more beneficial than additional examples. The improvement was primarily driven by higher precision with comparable recall, suggesting that the tags helped models focus on the correct lesion spans. Accordingly, we used lesion-tagged inputs throughout all LLM experiments. Generation was performed with settings intended to minimize stochastic variation (temperature = 0, top-$p$ = 1).

\begin{figure*}[!ht]
\centering
\begin{tcolorbox}[
  colback=gray!5,
  colframe=black!60,
  width=\textwidth,
  arc=3pt,
  boxrule=0.5pt,
  left=4pt,
  right=4pt,
  top=4pt,
  bottom=4pt]
\small
\textbf{Prompt for Incidentaloma Identification}

\begin{verbatim}
Role: You are a board-certified radiologist. 
Task: Analyze each report to verify incidentaloma status in target anatomies.

The list of lesions to consider is provided in the input using <LESION> </LESION> tags.

Exclusions:
Do not classify a lesion as an incidentaloma if:
- It is suspected or potential metastasis when the scan indication includes a known 
primary malignancy.
- It is found on surveillance scans (e.g., cirrhosis patients undergoing repeated liver 
imaging for HCC 
  or patients with known malignancy undergoing routine scans for metastases).
- It has been previously identified in a prior study.
- Its size change is mentioned ("stable", "decreased", "increased", "unchanged", etc.)
- Its clinical indication is related to the target lesion

Example output:
{
 'Lung Inci': {},
 'Liver Inci': {"LESION2":1},
 'Kidney Inci': {},
 'Adrenal Inci': {},
 'Pancreas Inci': {},
 'Thyroid Inci': {"LESION4":2},
 'Other Inci': {}
}

Empty dict: No incidentalomas
Category 1: Incidentalomas not requiring follow-up.
Category 2: Incidentalomas requiring follow-up.

Provide brief reasoning (<5 sentences) after JSON output.
\end{verbatim}
\end{tcolorbox}
\caption{Prompt used for verifying incidentaloma status across target anatomies. Exclusion criteria and examples are derived from the annotation guidelines.}
\label{fig:prompt}
\end{figure*}

Using these lesion-tagged radiology reports, we evaluated two prompt settings:
\begin{enumerate}
    \item \textbf{Base Prompt:} The lesion-tagged report is passed as-is.
    \item \textbf{With-anatomy Prompt:} The lesion-tagged report is supplemented with a concise mapping that pins each tag to previously extracted anatomy information described in \ref{sec:preprocessing}; the mapping is a single line such as \texttt{LESION1=Thyroid; LESION2=Pancreas; LESION3=Adrenal; \dots}. This pinning removes residual ambiguity about anatomy assignment while keeping all evidence in situ.
\end{enumerate}

\begin{figure*}[!ht]
    \centering
    \includegraphics[width=0.80\textwidth]{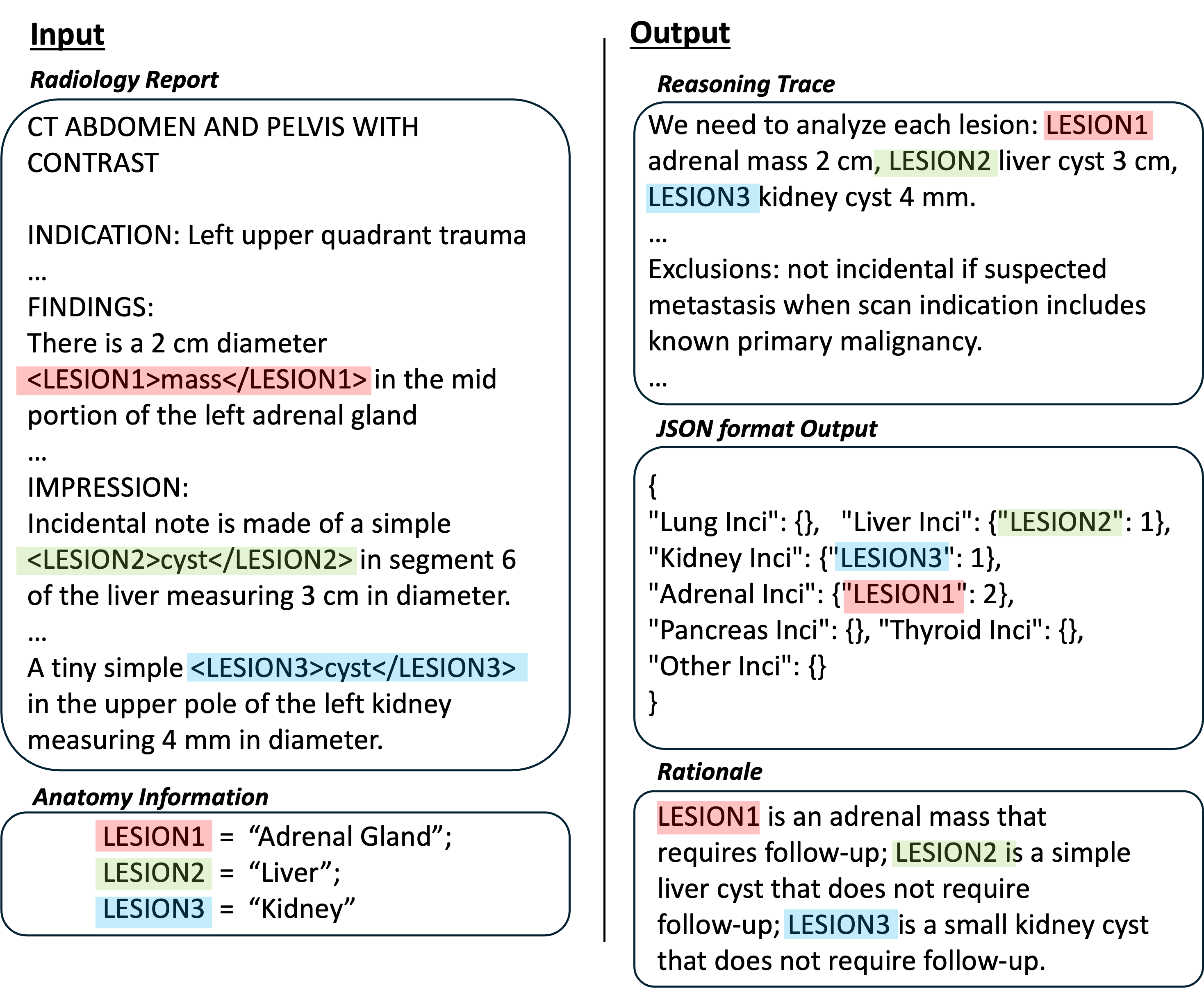} 
    \caption{Example of input and output used for LLM-based incidentaloma identification using GPT-OSS-20B. Lesions that are not returned in the JSON output are treated as No Incidentaloma (Class 0). Reasoning traces are available only in GPT-OSS-20B inferences, as GPT-4o does not provide reasoning trace outputs.
}
    \label{fig:llm_io_example}
\end{figure*}

An example of the With-anatomy prompt is shown in Figure~\ref{fig:llm_io_example}. Both prompting approaches use the same instruction described in Figure~\ref{fig:prompt} and the same report content, where lesions are tagged using numbered XML-format tags. In the With-anatomy approach, additional anatomical information is included. As illustrated in the LLM reasoning trace in Figure~\ref{fig:llm_io_example}, the model correctly attends to the target lesions by leveraging the numbered tags during inference.

We evaluated four generative LLM-based approaches under identical settings using the two prompts described above. The first was a LoRA (Low-Rank Adaptation) \cite{hu2022lora} fine-tuned Llama 3.1-8B model, adapted on our annotated radiology reports to generate incidentaloma labels in a domain-specific manner while keeping parameter updates lightweight. The second approach was a prompt-engineered version of Llama 3.1-8B, applied in a zero-shot fashion without gradient updates and evaluated in both prompt settings. The third approach employed GPT-4o, a proprietary state-of-the-art model, to benchmark performance against open-source alternatives. Finally, we evaluated GPT-OSS \cite{agarwal2025gpt}, an open-source GPT-style model that has demonstrated competitive performance across multiple public benchmarks. For computational feasibility, we used GPT-OSS-20B rather than GPT-OSS-120B, balancing inference time and resource requirements while maintaining strong performance. All experiments were conducted on our HIPAA-compliant secure server environment. Fine-tuning GPT-OSS-20B was not feasible because the model relies on generating explicit reasoning traces, which were not available for our clinical dataset. Fine-tuning GPT-4o was not performed to maintain consistency across proprietary models and to ensure a fair comparison within a prompt-only evaluation framework.

\subsection{Evaluation}
The held-out test set contains 80 double-annotated reports, yielding 560 anatomy-specific labels in total (seven anatomies per report). Of these labels, 50 are \textit{Incidentaloma--No Risk} (1) and 29 are \textit{Incidentaloma--Follow-up Required} (2); the remainder are \textit{No Incidentaloma} (0). The training set (n = 280) and validation set (n = 40) were used exclusively for prompt refinement in the LLM-based approaches and for model optimization in the supervised learning experiments.

Model performance was primarily evaluated using the F1-score, which balances sensitivity and precision. F1-scores were calculated for all three classes. To better capture performance on clinically meaningful cases, we additionally report an \textit{incidentaloma macro-F1}, defined as the mean of the F1 values for the two incidentaloma classes (1 and 2). Because the \textit{No Incidentaloma} class comprised the majority of samples and consistently yielded high performance across models, the incidentaloma macro-F1 offers a more balanced assessment focused on cases where incidentalomas are present.

\ifsubfile
\bibliography{mybib}
\fi

\section{Results}

\subsection{Performance Comparison}

As shown in Table~\ref{tab:results}, the best overall performance was achieved by the GPT-OSS-20b (With Anatomy) model, which obtained the highest F1-scores across all labels and an incidentaloma-positive macro-F1 of 0.79. GPT-4o (With Anatomy) was the second-best performer, with F1 scores of 0.82 and 0.71 for the incidentaloma-positive classes. These two anatomy-informed methods consistently outperformed all other systems, indicating that explicit anatomy context substantially improves incidentaloma classification.

Among supervised encoders, BioClinicalModernBERT (w/o CS) and ModernBERT (CS) both reached an incidentaloma macro-F1 of 0.70, with BioClinicalModernBERT performing better for Class 1 (0.79 F1) and ModernBERT slightly stronger for Class 2 (0.63 F1). Cost-sensitive (CS) learning produced only modest gains and mainly increased recall for minority classes.

Comparing model families, LLMs showed clear gains over supervised encoders. Only Llama 3.1-8B was fine-tuned using LoRA; all other LLMs, including GPT-4o and GPT-OSS-20B, were evaluated in a prompt-only configuration. GPT-4o (Base) already matched the strongest non-LLM baselines, and adding anatomical context further improved performance, yielding up to a $\Delta$+0.08 increase in macro-F1 (classes 1–2) and clearly surpassing all non-LLM methods. The consistent benefit of anatomy-aware prompting across both Llama and GPT-based architectures highlights the value of anatomy-aware contextualization of lesion findings in improving incidentaloma detection.

\begin{table}[ht]
\centering
\small
\setlength{\tabcolsep}{6pt}
\renewcommand{\arraystretch}{1.0}
\caption{Performance comparison of supervised encoders (with and without cost-sensitive learning (CS)) and LLM-based approaches on incidentaloma classification. F1 values are reported for each class (0: No Incidentaloma, 1: Incidentaloma--No Risk, 2: Incidentaloma--Follow-up Required). Overall Accuracy and Incidentaloma Macro-F1 (computed on Class 1 and 2 only) are also reported to show each model’s overall correctness and ability to detect and correctly classify incidentaloma-positive cases. Best values for each category are in bold.}
\label{tab:results}
\resizebox{\textwidth}{!}{%
\begin{tabular}{lccccc}
\hline
\textbf{Model} & \textbf{Class 0} & \textbf{Class 1} & \textbf{Class 2} & \textbf{Accuracy} & \textbf{Incidentaloma Macro-F1} \\
\hline

Inter-annotator Agreement (IAA) & 0.93 & 0.81 & 0.70 & 0.89 & 0.76 \\
\hline
\multicolumn{6}{l}{\textit{Supervised Encoder-based Models}} \\
\hline
BioClinicalModernBERT (w/o CS) & 0.99 & \textbf{0.79} & 0.61 & 0.95 & 0.70 \\
BioClinicalModernBERT (CS) & 0.99 & 0.72 & 0.60 & 0.95 & 0.66 \\
ModernBERT (w/o CS) & 0.99 & 0.76 & 0.60 & 0.95 & 0.68 \\
ModernBERT (CS) & \textbf{0.99} & 0.77 & \textbf{0.63} &\textbf{ 0.95} & \textbf{0.70} \\
\midrule
\multicolumn{6}{l}{\textit{Large Language Models}} \\
\hline
Fine-tuned Llama 3.1-8B & 0.96 & 0.62 & 0.46 & 0.91 & 0.54 \\
Llama 3.1-8B (Base) & 0.89 & 0.59 & 0.46 & 0.81 & 0.52 \\
Llama 3.1-8B (With Anatomy) & 0.90 & 0.61 & 0.64 & 0.82 & 0.63 \\
GPT-4o (Base) & 0.96 & 0.76 & 0.62 & 0.92 & 0.69 \\
GPT-4o (With Anatomy) & 0.97 & 0.82 & 0.71 & 0.94 & 0.77 \\
GPT-OSS-20b (Base) & 0.96 & 0.81 & 0.71 & 0.93 & 0.76 \\
GPT-OSS-20b (With Anatomy) & \textbf{0.97} & \textbf{0.84} & \textbf{0.73} & \textbf{0.94} & \textbf{0.79} \\
\bottomrule
\end{tabular}%
}
\end{table}

\subsection{Pairwise Statistical Significance Testing of Models on Incidentaloma-Positive Lesions}

\begin{figure}[!ht]
    \centering
    \includegraphics[width=\textwidth]{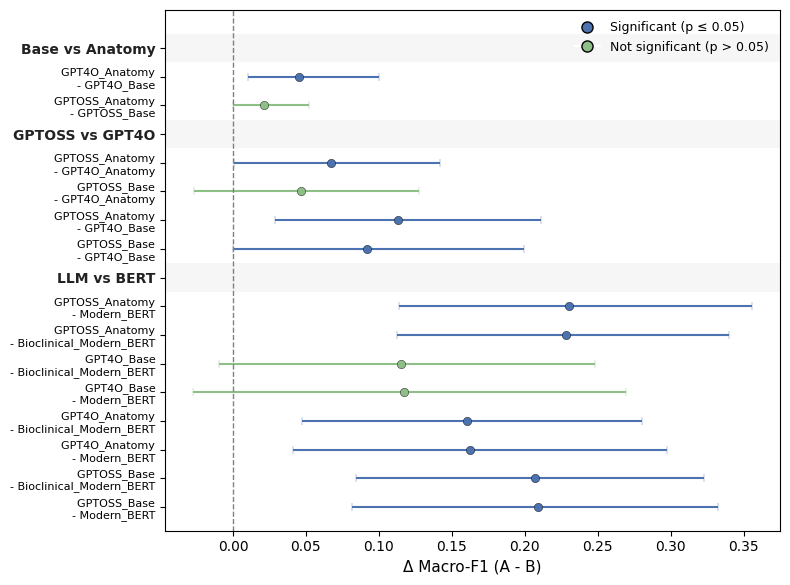}
    \caption{Pairwise non-parametric bootstrap comparison of model performance on incidentaloma-positive lesions. Each point represents the mean difference in Macro-F1 (A--B) across 1,000 lesion-level bootstrap samples, with horizontal bars showing 95\% confidence intervals. Comparisons to the right of zero indicate that Model A outperformed Model B.}
    \label{fig:pairwise_incidentaloma_forest}
\end{figure}

To evaluate statistical significance in model performance, we conducted a lesion-level bootstrap analysis restricted to lesions annotated as incidentalomas. This approach quantifies variability at the lesion level and focuses on clinically meaningful cases. Figure \ref{fig:pairwise_incidentaloma_forest} summarizes these comparisons, with each horizontal line representing the estimated CI for $\Delta$ Macro-F1. Llama-based models were excluded because of their consistently lower performance across all metrics.

Overall, both GPT-OSS (Base) and GPT-OSS (Anatomy) achieved higher Macro-F1 values than the supervised baselines (ModernBERT, BioclinicalModernBERT). GPT-OSS (Anatomy) provided the best overall performance, with confidence intervals lying entirely above zero when contrasted with all supervised encoders ($p < 0.01$). All GPT-based models outperformed the supervised methods by a statistically significant margin, underscoring the advantage of LLMs for lesion-level classification.

Consistent with the previous section, both GPT-OSS and GPT-4o showed significant gains when anatomical context was incorporated through anatomy-aware prompting. GPT-4o (Anatomy) significantly outperformed its base version ($p = 0.012$), and GPT-OSS (Anatomy) also improved over GPT-OSS (Base). These results demonstrate that anatomy-aware prompting improves robustness and clinical reliability of LLM-based systems for incidentaloma classification and is a generalizable strategy for radiology-specific information extraction.

\subsection{Error Analysis}

\subsubsection{Errors in Supervised Models}

We examined error patterns for BioClinicalModernBERT and ModernBERT, both with and without cost-sensitive (CS) training, to understand the limitations of supervised encoders. Clinical Longformer was excluded from detailed analysis because of its poor recall for follow-up-required cases and inconsistent error behavior.

BioClinicalModernBERT showed the most balanced performance among the supervised models but still underestimated lesion severity, frequently labeling follow-up-required incidentalomas as low-risk. Cost-sensitive training reduced some of these misses but increased false positives for benign lesions. ModernBERT showed a similar trend with generally lower recall and precision, often missing follow-up recommendations that appeared only in impression sections or in abbreviated form (e.g., ``f/u CT advised'').

Qualitative inspection indicated that both models struggled with hedged phrasing (e.g., ``likely benign,'' ``probably cystic''), complex size descriptions, and multi-lesion narratives mixing benign and higher-risk lesions. ModernBERT particularly struggled when follow-up recommendations were expressed indirectly or outside the main lesion description. LIME-based interpretability analyses \cite{ribeiro2016should} supported these findings, showing that token-level attributions were dominated by explicit lesion descriptors and common modifiers (e.g., ``subcentimeter,'' ``stable,'' ``mass''), while contextual indicators of clinical intent were underweighted.

\subsubsection{Errors in Generative LLMs}

The most critical errors involved missed incidentalomas, which represent potentially high-risk findings. GPT-4o (Base) missed 10 of 29 (34.5\%) such cases, GPT-4o (Anatomy) reduced this to 7 of 29 (24.1\%), GPT-OSS (Base) to 6 of 29 (20.7\%), and GPT-OSS (Anatomy) to 5 of 29 (17.2\%). These trends indicate that anatomy-aware prompting improves focus on lesion-specific information and reduces clinically important misses.

False positives, where normal lesions were misclassified as incidentalomas (Class 0\,→\,Class 1 or 2), were uncommon (3–6\%). Anatomy inclusion slightly reduced these rates for GPT-4o (4.8\%\,→\,3.5\%) and GPT-OSS (5.6\%\,→\,5.4\%). None of the models predicted no-risk incidentalomas as requiring follow-up. Across all models, a recurring pattern was underestimation of severity, where follow-up-required lesions were misclassified as no-risk (Class 2\,→\,Class 1). GPT-4o showed such underestimation in 13.8\% of error cases, while GPT-OSS reduced this to 10.3\%. These errors often appeared in reports with equivocal language or indirectly expressed follow-up intent. GPT-OSS (Anatomy) achieved the best balance, minimizing hazardous misses while keeping false positives low. 

To further characterize GPT-OSS (Anatomy), we reviewed its 33 errors among 560 evaluated lesions. Errors occurred across multiple anatomical sites (lung, liver, kidney, thyroid), indicating broader reasoning challenges rather than anatomy-specific effects. Three dominant mismatch types were identified:

\textit{Follow-Up Required vs. No-Risk (Gold = Class 2 → Model = Class 1, n=3)}\\
GPT-OSS (Anatomy) correctly identified the lesion but underestimated its clinical significance. These cases involved conditional language such as \textit{``tiny sub-6 mm lung nodule, follow-up if high risk''} or \textit{``indeterminate hypodensity on unenhanced liver CT''}, where reassuring phrases were overemphasized and qualifiers related to higher-risk populations or incomplete characterization were underweighted.

\textit{Incidentaloma Missed (Gold = Class 1 or 2 → Model = Class 0, n = 4)}\\
Four lesions labeled as incidentalomas were classified as non-incidentalomas because annotators had overlooked contextual cues such as ``stable'' or ``compared to prior exam,'' indicating prior evaluation. These were annotation errors rather than model failures. GPT-OSS (Anatomy) correctly identified all four as non-incidental, suggesting sensitivity to subtle contextual indicators that human annotators sometimes miss.

\textit{False Incidentaloma Detection (Gold = 0 → Model = 1 or 2, n = 26)}\\
False positives were the largest error category. GPT-OSS (Anatomy) misclassified 26 non-incidental cases as incidentalomas, split between no-risk (13) and follow-up-required (13). Most arose from differences in interpreting the relationship between the clinical indication and the lesion. For example, with clinical indication \textit{``previous CXR abnormal''} and \textit{``small opacity in the left lower lung''}, annotators linked the opacity to the prior abnormal CXR and labeled it non-incidental, whereas the model treated it as a separate finding requiring follow-up. In another case with indication \textit{``restaging''}, human annotators considered a lesion in a different anatomical region incidental, while the model interpreted the context as evidence of existing malignancy and classified it as non-incidental. These examples highlight how divergent interpretations of clinical intent and anatomical context can lead to disagreement between human experts and LLMs.

\ifsubfile
\bibliography{mybib}
\fi

\section{Discussion}

\subsection{Ensemble Effects and Majority-Vote Performance}

Twelve lesions were correctly classified by all GPT-based models but misclassified by both BERT-based models, whereas the opposite occurred only five times. Most of the GPT-correct and BERT-incorrect cases contained nodule management guideline templates that are automatically inserted into reports by the radiology information system (RIS) macros (e.g., \textit{``follow-up per Fleischner Society guidelines''}). These standardized template insertions often appear without explicit contextual linkage to an active finding, which likely caused confusion for the supervised models that rely heavily on local lexical cues rather than broader semantic context. In contrast, GPT models demonstrated stronger contextual reasoning, correctly linking each recommendation to its corresponding lesion finding. While these results highlight the contextual adaptability of LLMs, the BERT-based models showed a more conservative bias that often missed positive incidentaloma cases but produced fewer false positives, reflecting a precision-oriented decision boundary.

To integrate the complementary strengths of different modeling approaches, a simple majority-vote ensemble approach was constructed across six strong systems: GPT-4o (Base and Anatomy), GPT-OSS (Base and Anatomy), ModernBERT with cost-sensitive (CS) training, and BioClinicalModernBERT without CS. Each lesion’s final label was determined by the most frequent prediction across models. In case of ties during majority voting, the lowest numerical label was chosen (e.g., $0 < 1 < 2$), providing a conservative bias toward non-incidentaloma predictions. This parameter-free ensemble achieved the highest overall performance, yielding a Macro-F1 of 0.902 and surpassing all individual models. The resulting lesion-level F1 scores were well balanced across all classes (No Incidentaloma = 0.988, No Risk = 0.889, Follow-up Required = 0.830), representing consistent gains over the best individual model, GPT-OSS (With Anatomy), which achieved 0.968, 0.842, and 0.727 for the same categories, respectively. The ensemble therefore improved each class by a notable margin, particularly for identifying incidentalomas requiring follow-up, where the F1 value increased by more than 0.10. This indicates that combining the contextual reasoning of LLMs with the structured precision of supervised encoders may enhance reliability in clinical information extraction from radiology reports, particularly in cases involving ambiguous or uncertain phrasing.

While the ensemble achieved clear gains in robustness and overall accuracy, this approach also introduces practical trade-offs. Deploying multiple large-scale models in parallel increases computational overhead, inference latency, and system complexity, which may limit real-world scalability. Furthermore, ensembling can obscure the interpretability of individual model decisions, complicating clinical validation and downstream error analysis. Future implementations should therefore balance the benefits of ensemble stability against operational efficiency and transparency, particularly when integrating LLM-based systems into clinical workflows.

\subsection{Clinical Implications and Potential Applications}

The proposed incidentaloma identification approach has several potential applications for clinical decision support (CDS), workflow optimization, and follow-up tracking in radiology practice. It could be integrated at the point of order entry within CDS systems to automatically identify patients with a prior incidentaloma and recommend the most appropriate follow-up imaging study. When incorporated into radiology reporting software, the model could provide real-time guidance based on lesion size, imaging characteristics, and patient demographics, suggesting the most relevant clinical practice guideline for follow-up. In addition, the structured lesion-level outputs could be used to populate follow-up tracking systems, enabling longitudinal monitoring of incidental findings and improving adherence to recommended follow-up intervals. Moreover, it forms a foundation for evaluation about the downstream clinical and economic impact of imaging incidentalomas in the setting of patient co-morbidities and other lesions.

Our study also offers opportunities for improved explainability and transparency in clinical AI systems. Lesion-level analyses and LLM-generated explanations could be integrated into interactive dashboards, enabling radiologists to visualize model reasoning and validate follow-up recommendations. Such interpretability features would enhance clinician trust and support safe integration of AI-assisted tools into routine radiology workflows.

Collectively, these applications illustrate how automated incidentaloma detection can extend beyond research evaluation to support safer, more consistent, and guideline-concordant imaging care.

\subsection{Limitations}

Despite these findings, several limitations warrant consideration. First, the annotated dataset, while carefully curated and reviewed by radiologists, remains limited in size due to the intensive nature of manual lesion-level annotation. This constraint may limit the generalizability of results to less common anatomies or atypical phrasing styles. Second, radiology reporting conventions and imaging modalities vary widely across institutions, which could reduce the transferability of model performance to other health systems without additional adaptation or domain-specific tuning. Third, the evaluation primarily focused on text-based single-report analysis and did not incorporate temporal context from prior studies or longitudinal patient histories that could refine incidentaloma characterization and follow-up reasoning.

In addition, annotation subjectivity may have introduced minor inconsistencies, particularly in borderline cases where incidentaloma status or follow-up need was open to interpretation. While annotation review and consensus procedures were employed, subtle differences in annotator judgment could influence ground-truth labels. Furthermore, although model explainability was partially examined through token-level attribution and qualitative inspection, deeper interpretability analyses involving radiology experts are needed to fully understand the decision-making behavior of llms in complex radiology narratives.

Future work should therefore explore the integration of longitudinal patient data, cross-institutional validation, and more scalable annotation strategies, alongside improved interpretability frameworks, to enhance both the reliability and clinical trustworthiness of AI-assisted incidentaloma detection systems.

\ifsubfile
\bibliography{mybib}
\fi

\section{Conclusions}

This study presents a comprehensive evaluation of supervised transformer-based encoders and generative LLMs for automated identification and classification of incidentalomas in radiology reports. By introducing lesion-tagged inputs and anatomy-aware prompting, we demonstrate that generative LLMs, particularly GPT-OSS (With Anatomy), achieve substantially stronger and more balanced performance than conventional supervised encoders. These gains were most pronounced in incidentaloma-positive cases, where explicit lesion and anatomical context enabled more precise reasoning about clinical relevance and follow-up necessity.

Beyond quantitative performance, this study emphasizes a detailed lesion-level error analysis. Incorporating structured lesion cues, such as lesion tags and anatomical information, promoted more consistent lesion-level inference in LLMs, while transformer-based supervised models produced fewer false positives, reflecting a precision-oriented bias. Ensemble analysis further demonstrated that combining generative and encoder-based systems through majority voting stabilized predictions across clinically meaningful categories.

Although this work was conducted on data from a single institution and limited to individual reports without longitudinal context, the results provide clear evidence that structured lesion context and anatomical grounding are key to reliable and transparent LLM-based clinical reasoning in radiology. Future research should extend these findings through multi-institutional validation, temporal modeling, and incorporation of human-in-the-loop frameworks to ensure clinical robustness and generalizability.

Overall, this work advances the development of clinically aligned, anatomy-aware LLM frameworks for radiology report understanding, bridging the gap between automated text classification and clinical decision support in real-world radiology workflows.

\ifsubfile
\bibliography{mybib}
\fi

\section*{Acknowledgments}
This work was supported in part by the National Institutes of Health and the National Cancer Institute (NCI) (Grant Nr. 1R01CA248422-01A1)). The content is solely the responsibility of the authors and does not necessarily represent the official views of the National Institutes of Health.


\bibliography{mybib}


\end{document}